\documentclass[10pt,twocolumn,letterpaper]{article}

\usepackage{cvpr}
\usepackage{times}
\usepackage{epsfig}
\usepackage{graphicx}
\usepackage{amsmath}
\usepackage{amssymb}

\usepackage[pagebackref=true,breaklinks=true,letterpaper=true,colorlinks,bookmarks=false]{hyperref}

 \cvprfinalcopy % *** Uncomment this line for the final submission

\def\cvprPaperID{4291} % *** Enter the CVPR Paper ID here

\usepackage[utf8]{inputenc} % allow utf-8 input
\usepackage[T1]{fontenc}    % use 8-bit T1 fonts
\hypersetup{colorlinks=true,allcolors=black}
\usepackage{hyperref}       % hyperlinks
\usepackage{url}            % simple URL typesetting
\usepackage{booktabs}       % professional-quality tables
\usepackage{amsfonts}       % blackboard math symbols
\usepackage{nicefrac}       % compact symbols for 1/2, etc.
\usepackage{microtype}      % microtypography
\usepackage{graphicx}
\usepackage{amsmath}
\usepackage{amssymb}
\usepackage[dvipsnames]{xcolor}
\usepackage{wrapfig}
\usepackage{enumitem}
\usepackage{multirow}

\graphicspath{{./figures/}}

\newcommand{\name} {RF-ReID}

\newenvironment{Itemize}%
{\begin{itemize}%
\setlength{\itemsep}{0pt}%
\setlength{\topsep}{0pt}%
\setlength{\partopsep}{0pt}%
\setlength{\parskip}{0pt}}%
{\end{itemize}}

\ifcvprfinal\fi
\begin{document}
\title{Learning Longterm Representations for Person Re-Identification\\ Using Radio Signals}

\author{Lijie Fan\thanks{Indicates equal contribution.} \quad Tianhong Li$^{\ast}$ \quad Rongyao Fang$^{\ast}$ \quad Rumen Hristov \quad Yuan Yuan \quad Dina Katabi \\ MIT CSAIL}
	\maketitle

\begin{abstract}
	Person Re-Identification (ReID) aims to recognize a person-of-interest across different places and times. Existing ReID methods rely on images or videos collected using RGB cameras. They extract appearance features like clothes, shoes, hair, etc. Such features, however, can change drastically from one day to the next, leading to inability to identify people over extended time periods.  
In this paper, we introduce RF-ReID, a novel approach that harnesses radio frequency (RF) signals for longterm person ReID. RF signals traverse clothes and reflect off the human body; thus they can be used to extract more persistent human-identifying features like body size and shape. We evaluate the performance of RF-ReID on longitudinal datasets that span days and weeks, where the person may wear different clothes across days. Our experiments demonstrate that RF-ReID outperforms state-of-the-art RGB-based ReID approaches for long term person ReID. Our results also reveal two interesting features:  First since RF signals work in the presence of occlusions and poor lighting, RF-ReID allows for person ReID in such scenarios. Second, unlike photos and videos which reveal personal and private information, RF signals are more privacy-preserving, and hence can help extend person ReID to privacy-concerned domains, like healthcare.

\end{abstract}
\section{Introduction}

Person re-identification (ReID) aims to match a person-of-interest across different cameras, and at different times and locations. It has broad applications in city planning, smart surveillance, safety monitoring, etc. It is challenging because the visual appearance of a person across cameras can change dramatically due to changes in illumination, background, camera view-angle, and human pose. With the success of deep learning, several ReID models \cite{ejaz2015deepreid2,wei2014deepreid,shizhe2016deeprank,weihua2017duadruplet,alexander2017triplet,yifan2018PCB,gao2018revisiting,zhun2018style-adapt} have managed to extract appearance features that are view-invariant  across cameras, leading to good performance on various person ReID datasets \cite{chang2018multi}. 

However, another great challenge has rarely been investigated: human visual appearance can also change drastically over time. For example, people may revisit the same shop on different days wearing different clothes and hair styles, and a thief can deliberately change his clothes to mislead the surveillance system. A robust person ReID system should be able to match people despite appearance changes. Unfortunately, existing RGB-based person ReID systems have severe limitations in achieving this goal since they intrinsically rely on appearance information such as clothes, shoes, hair, bags, etc. \cite{yifan2018PCB,hantao2019partloss,liming2017partaligned}. All these features are short-lived and can become ineffective the next day. To achieve robust person ReID, the system should be able to extract longterm identifying features that persist for weeks and months.

%Person Re-identification (ReID) is an important task in computer vision which aims to find out the person-of-interest across different cameras. It has broad applications in people's everyday life, such as video surveillance and safety monitoring. In the past several years, the success of Deep Learning and Convolutional Neural Networks led to great development in Person ReID \cite{ejaz2015deepreid2,wei2014deepreid,shizhe2016deeprank,weihua2017duadruplet,alexander2017triplet,yifan2018PCB,gao2018revisiting,zhun2018style-adapt}. They managed to extract view-invariant visual appearance characteristics across cameras and achieve superior performance on many person ReID datasets.

%However, extracting only view-invariant appearance features is far from enough for robust person ReID. A robust person ReID system should not only be able to match people across multiple cameras, but also capable of matching people who revisit the same place at different time. This is important in many real world applications, including surveillance where the person-of-interest can change clothes and appearance, or customer behavior analysis where the same person can revisit the shop wearing different clothes in different days.

\begin{figure}[t]
\begin{center}
\includegraphics[width=1.0\linewidth]{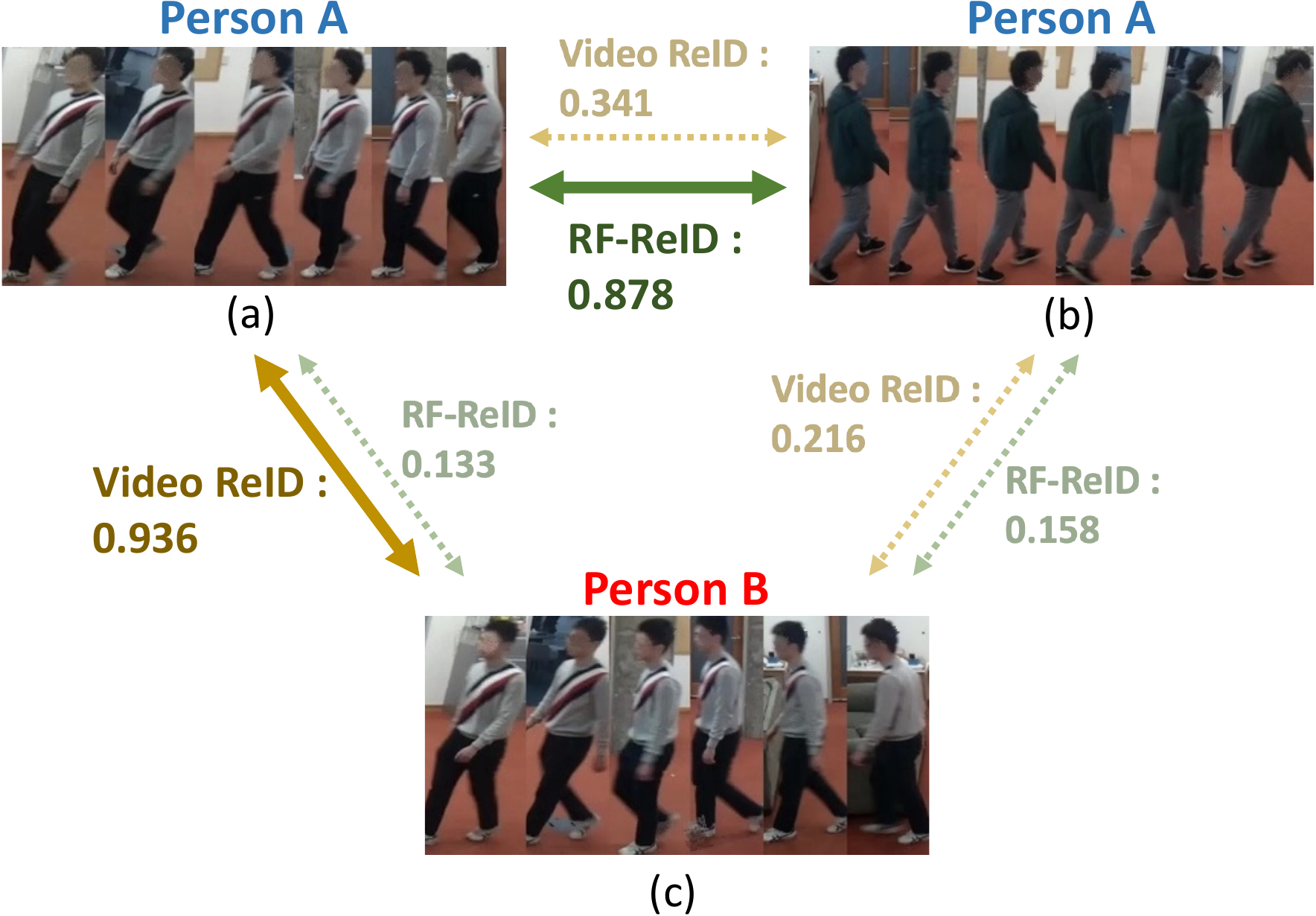}
\end{center}
\vspace{-10pt}
\caption{\footnotesize{Similarity score between people as computed by RF-ReID and a state-of-the art video-based ReID model (the larger the score, the higher the similarity). (a) and (b) show the same person wearing different clothes, and (c) shows a different person wearing the same pullover as the top-left person. The video-based ReID model relies on appearance features and thus wrongly predicts (a) to be close to (c), while RF-ReID captures body shape and walking style and can correctly associate (a) with (b).}}\label{fig:teaser}
\vspace{-18pt}
\end{figure}

But how can we capture persistent features suitable for person ReID? Wireless signals present a good solution. Radio frequency (RF) signals in the Wi-Fi frequency range  traverse clothes and reflect off the human body. 
Unlike cameras, wireless signals could extract intrinsic features of the human body, such as body size or shape. These features are relatively stable over days and months, enabling a more robust longterm ReID system. Moreover, previous works have shown the possibility of tracking people's 3D skeletons and walking patterns using RF signals, which can be used as longterm identifying features \cite{zhao2019through,li2019making,zhao2018rf-based,zhao2018throughwall, nambiar2019gait}. Furthermore, unlike RGB-based person ReID methods which struggle in the presence of occlusion and poor lighting, RF signals traverse walls and can enable ReID through occlusions and in dark settings.

However, applying RF signals to person ReID presents several challenges.  First, training an end-to-end model only using people's IDs as labels  leads to overfitting problems. ID labels provide a rather weak supervision. The model will tend to learn environment-dependent short-cuts such as the position a person usually stays at. Second, unlike RGB images, a single RF snapshot contains reflections from few body parts, and misses the rest of the body. This is due to a special property called specularity~\cite{beckmann1987scattering}. The human body acts like a mirror for RF signals in the Wi-Fi range, and signals reflected away from the receiver will not be captured. As a result, a single RF snapshot does not contain enough information about the whole body to identify the person.

To solve the overfitting problem, we propose a multi-task learning framework and an environment discriminator. Apart from predicting the identity of the person, we force features from our model to contain enough information to predict the 3D skeleton of the person. 
%This forces the model to learn information related to people. 
We further add an environment discriminator to force the features to be environment independent. This discriminator is co-trained with the feature extraction network in an adversarial way, making features from different environments indistinguishable by the discriminator. To solve the specularity of RF signals, we add a hierarchical attention module that effectively combines information regarding the human shape and walking style across time, i.e., across multiple RF snapshots. 

We introduce RF-ReID, an RF-based person ReID model that can extract longterm identifiable features, and work under occluded or inadequate lighting conditions. RF-ReID takes wireless signals as input, extracts identifiable features from the reflection of the human body, and identifies the person with the extracted feature. It performs robust person ReID across both time and space. \autoref{fig:teaser} shows an examples that demonstrates the effectiveness of RF-ReID. The same person wearing different clothes in (a) and (b)) is mistaken as a different individual by state-of-the-art video ReID~\cite{gao2018revisiting}, while two different people wearing the same clothes in (a) and (c) are wrongly identified to be the same person. In contrast, RF-ReID can accurately identify (a) and (b) to be the same person, and (a) and (c) to be different.

%Person A wearing 2 sets of different clothes ((a) and (b)) are mistakenly identified to be different by video ReID, but are correctly identified to be the same by RF-ReID. Person A and Person B wearing the same clothes ((a) and (c)) are mistakenly identifies to be the same by video ReID, but are correctly identified to be the different by RF-ReID. 

%To evaluate RF-ReID, we collect data using five radio devices from 5 different locations, and next to each radio device we deploy an RGB-camera to record corresponding videos for labeling. The dataset spans across 15 days and contains 100 different identities. Since our dataset is across days, it is common that a bunch of people in our datasets appear multiple times in different clothes, and different people may appear in similar clothes. Extensive experiments demonstrate that RF-ReID performs robust person ReID and outperforms the state-of-the-art RGB-based ReID approaches by a large margin. It also shows its ability to work through occlusions and in poor lighting conditions when the camera fails completely.

We evaluate RF-ReID on two datasets. {\bf (A) RRD-Campus:}  The first dataset is collected using five radios deployed in different locations on  our campus.  Ground truth ID labels are collected using RGB video cameras colocated with each radio. The resulting dataset contains 100 different IDs, and spans 15 days. People appear multiple times in different clothes and different people may wear similar clothes.   {\bf (B) RRD-Home:}  The second dataset was originally collected to assess the viability of tracking movements of Parkinson's patients in their homes using RF signals.  It includes data from 19 homes, and an average of one week per home. Ground truth IDs are obtained through manual labeling by comparing the movements from wearable accelerometers with movements from RF signals. More details about the datasets are available in \autoref{sec:dataset}.  
 
We train a single model to ReID people in both datasets. As described above we use a discriminator to ensure that the representation is environment independent. Since RRD-Campus includes colocated RGB videos, we compare our model with state-of-the-art video-based ReID\cite{gao2018revisiting}. The results show that RF-ReID outperforms the state-of-the-art RGB-based ReID approaches by a large margin. They also show RF-ReID ability to work through occlusions and in poor lighting conditions when the camera fails completely.  RF-ReID also works well on RRD-Home, which contains RF signals from the homes of Parkinson's patients.  This result shows that our method can ReID people in private settings like homes, locker rooms, and other private locations where one cannot deploy cameras. Interestingly, this result leads to a new concept -- privacy-conscious ReID, where people may be re-identified without capturing their personal information, e.g., pictures or audio clips.

To summarize, this paper makes two key contributions:
\begin{Itemize}
    \item First, it investigates the task of longterm person ReID, which identifies a person regardless of appearance changes over time.  It proposes a novel model that leverages RF signals to achieve longterm person ReID. It further demonstrates that the model is robust to occlusion and poor lighting.
    \item Second, it introduces the concept of privacy-conscious ReID as the ability to identify encounters with the same person without collecting personal or private data like pictures, videos, or audio clips.  The paper also demonstrates the first such privacy-conscious ReID model. 
\end{Itemize}

\section{Related Works}
\noindent\textbf{(a) RGB-based ReID. }
There are mainly two categories of RGB-based ReID: image-based and video-based. Early approaches for image-based ReID rely on hand-crafted features based on color descriptors, and optimize some distance metric relative to those descriptors~\cite{gong2014person,liao2015person,roth2014mahalanobis,lisanti2015person,vezzani2013people}. Early video-based ReID models use spatio-temporal descriptors like HOG3D \cite{klaser2008spatio} and gait energy image (GEI) \cite{han2005individual} to extract additional temporal information.

Recent approaches rely on deep learning and can be divided into two categories. The first category uses classification models similar to image or video classification tasks \cite{alex2012alexnet, carreira2017quo}. The second category uses siamese models which take a pair or triplet of images or videos as input, and uses pairwise or triplet loss to train the model~\cite{ejaz2015deepreid2,alexander2017triplet,weihua2017duadruplet}. Some Video-based ReID models further aggregate temporal information through temporal attention or RNN networks~\cite{gao2018revisiting, liu2017quality, mclaughlin2016recurrent, xu2017jointly, yan2016person}.

Both image-based and video-based ReID methods tend to extract short-lived features such as clothes, shoes, bags, hair styles, etc. \cite{chang2018multi} and hence struggle to re-identify a people across days and weeks.

\noindent\textbf{(b) RF-based Person Identification.}
Research in wireless systems has explored person identification using radio signals. Previous work can be divided into two categories. The first category uses the signals transmitted by portable devices (e.g., cellphones) to track and identify each person \cite{kotaru2015spotfi,xiong2013arraytrack,vasisht2016decimeter}. Such systems require a person to wear or carry sensors, which limits their utility and robustness. The second category analyses signal reflections off people's bodies to identify each person \cite{hsu2019enabling,adib2015smart,hsu2017zero,hsu2017extracting,pu2013whole,rahman2015dopplesleep,wang2015understanding,wang2014eyes,hong2016wfid,shi2017smart,wang2016gait,zeng2016wiwho}. Past systems however classify  small number of people (<10) in the same environment, cannot generalize to new identities unseen in the training set, and typically require the person to walk on certain constrained paths~\cite{wang2016gait, zeng2016wiwho}. In contrast to all past work, we are the first to achieve person ReID in the wild with RF signals, and without requiring people to wear sensors or move on specific paths. Furthermore, our model generalizes to new people and new environments unseen during training. 

%Previous RF analysis works related to identifying humans either (1) Require additional sensors \cite{hsu2019enabling, klaser2008spatio, xiong2013arraytrack, vasisht2016decimeter}; or (2) Focus on person classification rather than person ReID \cite{hong2016wfid, hsu2019enabling}, they only classify a few people ($\leq10$) in the same environment, and can't generalize to new people; or (3) Require the person to walk in certain constrained paths \cite{wang2016gait, zeng2016wiwho}. Hence none of these methods work for our dataset.  We are the first to achieve person ReID in the wild with RF signals, and without requiring people to carry/wear sensors or move on specific paths. 
\section{Radio Frequency Signals Primer}\label{sec:rf}
\begin{figure}[t]
\centering
\includegraphics[width=0.8\linewidth]{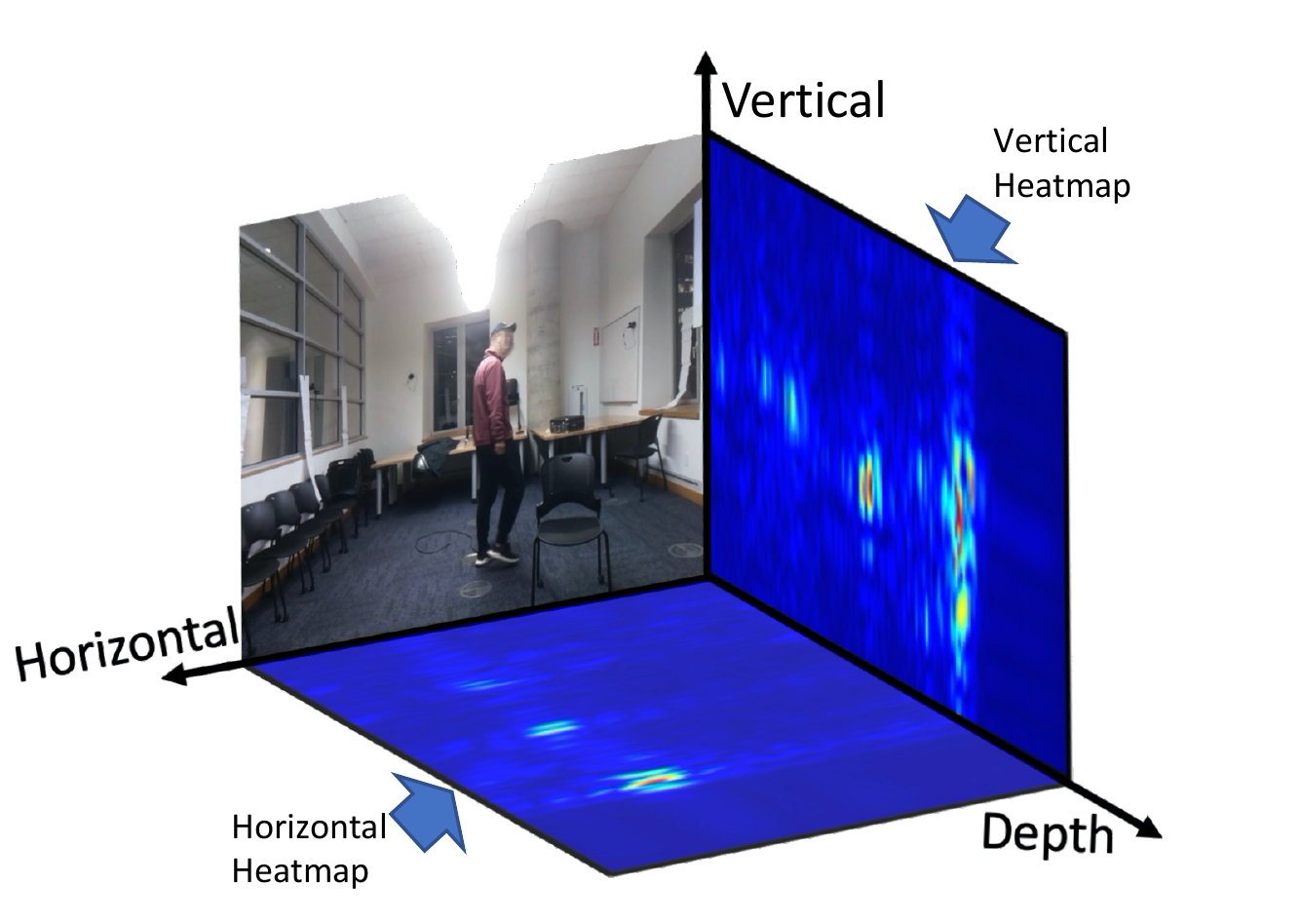}
\caption{\footnotesize{RF heatmaps and an RGB image recorded at the same time.}}	\label{fig:rf-heatmaps}
\vspace{-15pt}
\end{figure}

We use an FMCW radio widely used in previous work on RF-based human sensing \cite{zhao2018rf-based, lien2016soli, zhang2018latern, chetty2017low, peng2016fmcw, tian2018rf, hsu2019enabling, zhao2016emotion, zhao2017learning, zhao2019through, li2019making}.  The radio is equipped with two antenna arrays: horizontal and vertical. It operates between 5.4 and 7.2~GHz and can sense people up to 12m away from the device. 

\textbf{(a)~~RF Heatmaps}:  The RF signal at the output of the radio takes the format of two 2D heatmaps: one from the horizontal array and the other from the vertical array, as shown in \autoref{fig:rf-heatmaps} (red refers to large values while blue refers to small values). The horizontal heatmap is a projection of RF signals on the plane parallel to the ground, and the vertical heatmap is a projection of RF signals on a plane perpendicular to the ground. Intuitively, we can treat these heatmaps as depth maps, where higher values correspond to higher strength of signal reflections from a location. The radio generates 30 horizontal-vertical heatmap pair per second; we call each pair an RF frame. 

\autoref{fig:rf-heatmaps} reveals that RF signals have different properties from vision data. The human body is specular in our frequency range  \cite{beckmann1987scattering}. RF specularity occurs when the wavelength of the signal is larger than the roughness of the object's surface. In this case, the object acts like a mirror as opposed to a scatterer.  The signal from each body part may be reflected towards our sensor or away from it depending on the orientation. Therefore, each RF frame contains reflections from a subset of body parts, making it hard to obtain identifiable information from a single RF frame.  

\textbf{(b)~~RF Tracklets}: Prior work has demonstrated that RF signals can be used to detect, localize and track people \cite{adib20143d}. We use this technique to extract RF tracklets from the RF heatmaps.   As shown on the left of \autoref{fig:method}, a tracklet extracts from the horizontal and vertical heatmaps the RF signals reflected off a person, and the bounding box of that person at each time step (white box in figure).  Since one RF tracklet always corresponds to one person, the ReID task is performed across different RF tracklets.

\textbf{(c)~~Skeletons from RF}:  3D Human skeletons can be generated from RF signals using the approach in \cite{zhao2018rf-based}. The generated 3D skeleton data contains the 3D coordinates of $18$ major human body joints at each time step, as specified in \cite{fang2017alphapose}, which can be used to assist the task of person ReID.

\section{RF-ReID}

\begin{figure*}[htbp]
\begin{center}
\includegraphics[width=1.0\linewidth]{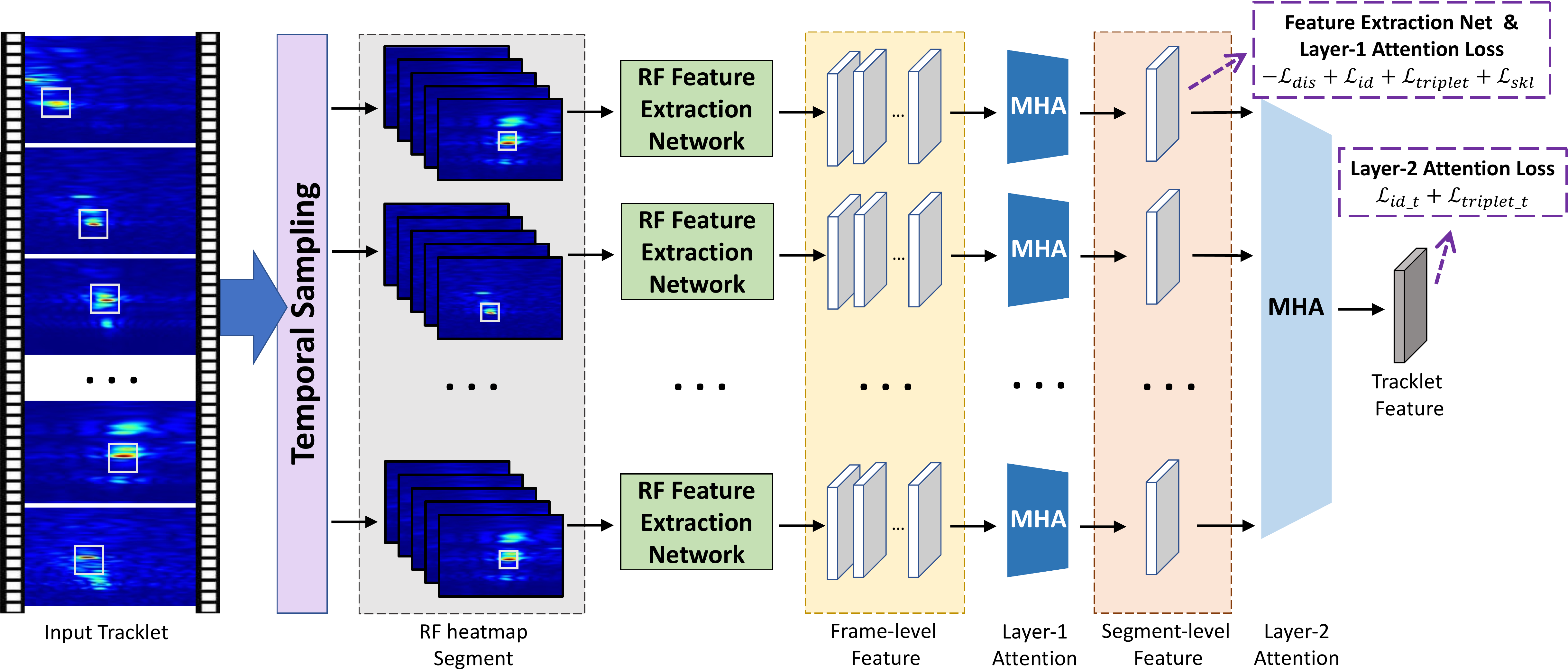}
\end{center}
\vspace{-10pt}
\caption{\footnotesize{Model architecture. \name\ takes an RF tracklet as input. It samples RF segments of 3 seconds (90 frames), and extracts frame-level features with an RF feature extraction network (shown in green). These features are processed by a multi-headed hierarchical attention module (i.e., MHA) with two sub-modules;
 the first attention sub-module (dark blue) extracts segment-level features and the second sub-module (light blue) extracts tracklet-level features. The supervision for training the RF feature extraction network and the first attention sub-module is added to the segment-level features, and the supervision for the second attention sub-module is added to the tracklet features.}} 
\label{fig:method}
\vspace{-10pt}
\end{figure*}

RF-ReID is an end-to-end model for person ReID using RF signals. As shown in \autoref{fig:method}, our model takes an RF tracklet as input. It then extracts features from the tracklet using an RF feature extraction network. It then aggregates temporal information through a learnable hierarchical attention module to generate a feature map. During training, these features are supervised in a multi-task learning manner using identity classification loss, triplet loss and skeleton loss. We further add an additional environment discriminator loss to force the model to learn environment-invariant features. This allows our model to generalize to new environments not seen during training.  Below, we explain each RF-ReID component in detail.

\subsection{RF Feature Extraction Network}
Since RF tracklets can have different durations, we first perform temporal sampling on each tracklet before extracting features from it. For each RF tracklet, we uniformly sample 25 segments from it, where each segment contains 3 seconds (90 frames) of RF heatmaps. 

We adopt an architecture similar to \cite{zhao2018rf-based} for our backbone feature extraction network. The network first uses spatio-temporal convolutions to extract global features from the input RF frames. We then crop out the region of interest around the tracklet trajectory in the feature map. Finally, the cropped features are fed into a sub-network to generate frame-level identifiable features for person ReID. 
%Since we only use one layer with stride 2 in the feature extraction network, the output features will have length 45 in its temporal dimension.

\subsection{Hierarchical Attention Module}
The RF feature extraction network generates relevant features from each RF segment, where a segment is a clip of 90 RF frames (3 seconds).  As mentioned in Section~\ref{sec:rf}, due to specularity, each RF frame contains information only about some body parts. So we need to aggregate features across frames in the same tracklet. 
%Simply performing average pooling over the features in a tracklet is not a desirable solution because it can lose information. 
To solve this problem, we propose a learnable two-step hierarchical attention module to aggregate information across each tracklet. 

There are two kinds of information in an RF tracklet that is relevant to person identification: shape and walking style. The coarse shape of the person can be obtained by aggregating information from several seconds of RF signals. This is because when a person moves, we can receive signals reflected from different body parts according to their orientation with respect to the radio. Thus, the first attention block is added on frame-level features to aggregate the shape information within each 90-frame segment (3 seconds).

The walking style, on the other hand, is a feature that can only be inferred from a longer temporal span.  However, within an RF tracklet, there can be many non-walking periods where the person may stop, stand by, sit down, tie their shoes, etc. Those periods cannot be used to infer the walking style. Therefore, we use the second attention block to attend to features from different segments across the tracklet and aggregate them to generate one final feature vector for each tracklet. 
%Moreover, since our multi-task learning can generate 3D skeletons of the person
%we also add the predicted skeleton as additional input to the second attention block for help.
%Both of the two attention block can be trained end-to-end together with the feature extraction network.

\subsection{Multi-task Learning for Identifiable Features}\label{sec:losses}
\begin{figure}[t]
\begin{center}
\includegraphics[width=1.0\linewidth]{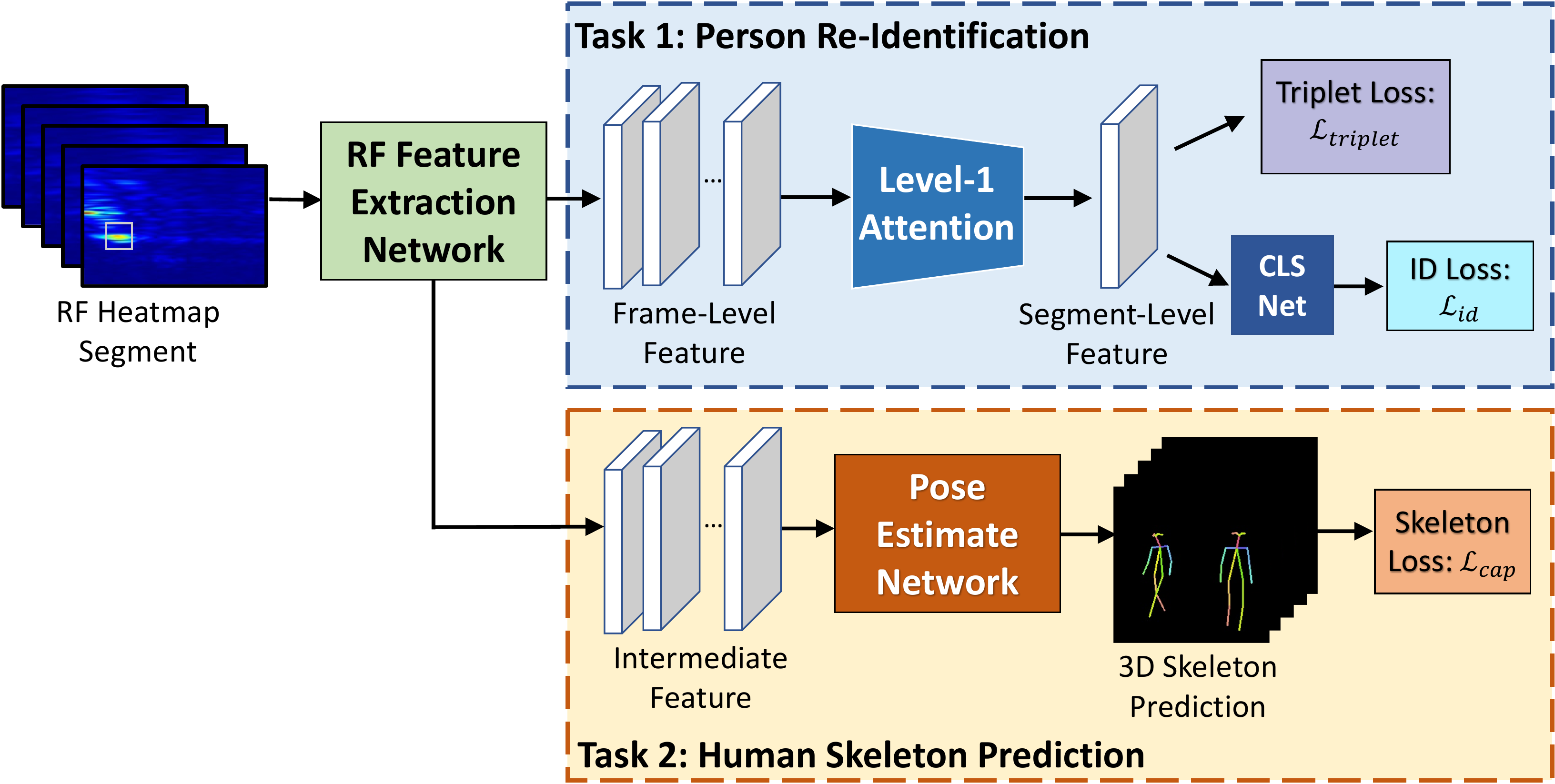}
\end{center}
\vspace{-10pt}
\caption{\footnotesize{Illustration of multitask learning in RF-ReID. The blue box corresponds to the the task of person re-identification, and the yellow box corresponds to the task of 3D skeleton prediction.}}
\label{fig:multitask}
\vspace{-15pt}
\end{figure}

%After extracting features from RF signal, 
%We need add supervision to train the RF feature extraction network so that the generated features can be used for person ReID. 
To train the RF feature extraction network, we add supervision to the segment-level features (orange box in \autoref{fig:method}).
As shown in \autoref{fig:multitask}, we first add two losses widely used in prior works on RGB-based person ReID: identification loss $\mathcal{L}_{id}$ and triplet loss $\mathcal{L}_{triplet}$. For the identification loss, the segment-level features are further passed through another classification network with two fully-connected layers to perform ID classification. This task helps the model learn human identifiable information from RF signals. The triplet loss \cite{alexander2017triplet} is computed as
$$\mathcal{L}_{triplet} = \max(d_p - d_n + \alpha, 0),$$ 
where $d_p$ and $d_n$ are the $L_2$ distances 
of segment-level features from the same person and features from different people, respectively. $\alpha$ is the margin of triplet loss ( $\alpha$ is set to 0.3). This loss enforces features from different people to be far away from each other and those from the same person to be close. 

The first attention layer can be trained end-to-end with the feature extraction network. For the second attention layer, we first generate features for each segment, then we use the second attention layer to aggregate them and train the second attention layer using the ID loss $\mathcal{L}_{id\_t}$ and triplet loss $\mathcal{L}_{triplet\_t}$ on the aggregated feature.

Besides these two losses, we force the model to learn to infer the person's skeleton from RF signals and add an additional skeleton loss $\mathcal{L}_{skl}$ for supervision. Specifically, we take the intermediate features in our feature extraction network two layers above the frame-level features and feed them into a pose-estimation sub-network that generates 3D human skeletons. The skeleton loss is a binary cross entropy loss similar to the one used in \cite{zhao2018rf-based}. This loss forces the features to contain enough information for skeleton generation, which can help the ReID task in capturing  person's height and walking style. The skeleton loss also acts as a regularizer on the extracted features to prevent overfitting.

\subsection{Environment Discriminator} 
RF tracklets can have identifiable patterns strongly related to the environment. For example, a person is much more likely to enter his or her own office than other people. As a result, the model can use such environmental features as shortcuts to identify people based on their paths. This will severely harm the model's ability to generalize across different environments or identify people when they do not follow their usual paths. For example, in our Campus Dataset, the model will have difficulty identifying a faculty who is visiting another faculty in their office. In our Home Dataset, the model will learn the specific path that each person walks in their home, and fail to generalize to new unseen homes. 

To solve this problem, we need to eliminate the environment-dependent factors during the training process. Thus, we consider signals from each radio location as one environment, and train a discriminator to predict the environment of the signal. The discriminator is trained in an adversarial way so that eventually the model will eliminate the features that are environment dependent. The discriminator operates on segment-level features as shown in \autoref{fig:method}. A cross entropy loss is used to train the discriminator to predict the environment. The discriminator loss is subtracted from  the loss of multi-task training in the feature extraction network. Denoting the RF feature extraction network as $F$ and the environment discriminator as $D$, the discriminator loss is:
\vspace{-10pt}
\begin{eqnarray*}
\mathcal{L}_{dis}&=&-\sum_{c=1}^M y_{c}\log(D(F(x))_c),
\end{eqnarray*}
where $x$ is the input tracklet, $M$ is the total number of environment and $y_c$ is the binary indicator of which environment the current tracklet belongs to.  The optimization target is:
\vspace{-5pt}
\begin{eqnarray*}
\min_{F} \max_{D}V(F,D)&=&-\mathcal{L}_{dis}+\mathcal{L}_{id}+\mathcal{L}_{triplet}+\mathcal{L}_{skl}
\end{eqnarray*}

%We train the discriminator for one step each time we train the feature extraction network for one step. 

\section{Dataset}
\label{sec:dataset}
%In order to train a model to identify people from RF signals, we need a dataset of RF signals and the corresponding identity labels. 
We use the following two datasets. 

\subsection{RRD-Campus} 
RRD-Campus refers to an RF-ReID Dataset collected on our Campus. It contains RF signals and synchronized RGB video data. The video data is used both for ground truth labeling and to compare with RGB-based video and image based ReID methods.  

The data is collected by deploying 5 radios in 5 different locations across our campus and collecting data for 15 days. Each radio is colocated with an RGB video camera.  We synchronize the video data and the RF signals using the NTP protocol, with a maximum time difference of 10 ms. 

\textbf{Labeling}:
%One reason that previous RGB based ReID models tend to learn longterm-variant appearance features is that the labeling process of the dataset is highly dependent on the appearance of people: the clothes they wear, the things they carry, the hairstyles they have, etc. In their labeling process, the labelers use different synchronized cameras next to each other to track people. They only label the person across a short period as the same identity, who is always wearing the same set of clothes. 
%To solve this problem and achieve the goal of longterm-invariant person ReID, 
We deploy our data collection system at places that tend to be revisited by the same people, such as the lounge area in front of a lab or a set of offices. We collect video with $960\times720$ resolution and $15$ FPS to assist in the labeling process.
We then use the video to label all people who show up repeatedly, even if they have different clothes. We further ask the people in the dataset to double-check their own data and make sure that all RF tracklets with their identity are labeled correctly. 
%These procedures ensure that RRD-Campus has high quality.

\textbf{Statistics}: The dataset contains $100$ identities in total. On average, each identity has $8.63$ RF tracklets, and each tracklet spans over $11.0$ seconds. People in the dataset may change clothes across different days, and our data collection system does not interfere with people's normal activities.
% and each identity in the test set has $1$ query tracklet and $7.55$ gallery tracklets.} 

\subsection{RRD-Home}
RRD-Home is based on the dataset in previous work on Parkinson's Disease analysis with RF signals~\cite{kabelac2019passive}. The dataset is collected by deploying RF devices in 19 different homes to obtain RF tracklets for moving people, where each home is inhibited by a Parkinson patient and a normal person (typically the spouse). The Parkinson patient is asked to wear an accelerometer to collect the acceleration data. The collected dataset contains the RF tracklets for the people in each home, and the corresponding accelerometer data for the patients. Note that the acceleration data is only used for labeling and is not an input to our model.

\textbf{Labeling}:
Accelerometers are widely used for in-home behavior analysis~\cite{hsu2019enabling}. In RRD-Home, We use accelerometer data to assist in the labeling process. In each home, we first calculate the location, moving distance and speed for each RF tracklet, and then associate the patient's accelerometer data with the corresponding RF tracklets based on the similarity of movement properties. RF tracklets whose motion is synced with the acceleration data from the wearable on the patient are labeled with the patient's ID, whereas the tracklets whose motion does not match the acceleration data are labeled with the ID of the other person in the home. (We ignore RF tacklets from periods during which there is no acceleration data.)
 
%The devices are deployed in people's home, like in bedroom, living room etc. We use accelerometer data to assist the labeling process, as it has repeat patterns when the person is moving. Specifically, In each home, we associate the accelerometer data with the RF Tracklets data, and label those trackelets with synced movements as acceleration data to be the patient, and label the other trackelets to be the other person.

\textbf{Statistics}: The dataset contains $38$ different identities in $19$ different homes.  The data spans 127 days with an average of one week per home.  Each identity has $165.91$ RF tracklets on average. Each RF tracklet spans over $9.24$ seconds. People in the dataset can change clothes across days, and the data collection system does not interfere with people's normal activities.
%Each identity in the training set has on average $165.91$ RF tracklets. For each identity in the test set, we sample $1$ query RF tracklet and $9.82$ gallery RF tracklets.

\subsection{RF-Based Skeletons} As mentioned in Section~\ref{sec:losses}, to assist training our ReID model we force it to learn features useful for inferring a person's skeleton from RF signals. This is done by leveraging the datasets from past work on 3D RF-based pose and action estimation~\cite{li2019making,zhao2018throughwall}.

\section{Experiments}
We evaluate the effectiveness and practicality of RF-ReID and compare it with image and video based ReID.

\subsection{Experimental Setup}
\textbf{Training \& Testing}: 
\textbf{(1) RRD-Campus.} 
We split RRD-Campus into a training set with $ 60 $ identities and a test set with the other $ 40 $ identities. As common in RGB-based ReID datasets, we randomly select one sample from each identity in the test set to build the query set and group the remaining samples as the gallery set. 
\textbf{(2) RRD-Home.} 
The training and test sets of RRD-Home contain $ 13 $ and $ 6 $ different homes, respectively. Each home has 2 different identities corresponding to the two inhabitants. The query set and the gallery set is constructed similarly to RRD-Campus. 
\textbf{(3) RRD.} We also combine RRD-Campus and RRD-Home to form a larger dataset RRD. The training set of RRD is the combination of RRD-Campus training set and RRD-Home training set. 
%We combine the training sets from  RRD-Campus and RRD-Home and train one RF-ReID model that works for all environments, and people in our datasets. 
%We test this model on each dataset separately, as well as on the combined 
%The evaluation is performed on the testing sets of RRD-Campus and RRD-Home separately to allow for comparing the difficulty of ReID on each dataset.  
We evaluate our model on both the individual datasets and the combined one.  We perform 5-fold cross-validation, where each time we randomly assign identities to the training set and test set.

%\textbf{Dataset}: To train and evaluate our model, we split our collected dataset into a training set and a testing set. The training set consists of $60$ identities and the testing set consists of $40$ identities different from the training set. As common in ReID datasets, we randomly select one sample (tracklet for RF-ReID and corresponding videos for RGB=based ReID) from each identity in testing set to build the query set. Other samples are then grouped as gallery set. 

\vskip 0.05in
\textbf{Evaluation Metrics:} 
During testing, the query samples and gallery samples are encoded to feature vectors using RF-ReID. Then we calculate the cosine distance between each query sample's features and each gallery sample's features and rank the distance to retrieve the top-N closest gallery samples for each query sample. We compute the standard evaluation metrics for person ReID based on the ranking results: mean average precision score (mAP) and the cumulative matching curve (CMC) at rank-1 and rank-5. 
%All results are reported as the average of 10 trials.

\vskip 0.05in
\textbf{RGB Baseline Models:}
To demonstrate the effectiveness of \name, we compare it with one state-of-the-art image-based person ReId model \cite{luo2019bag} and one state-of-the-art video-based model \cite{gao2018revisiting}. The comparison is performed on RRD-Campus since only RRD-Campus is collected with synchronized RGB videos. We first train the image-based and video-based person ReID models on the commonly used Market1501 \cite{zheng2015market} and MARS \cite{zheng2016mars}, respectively. We then fine-tune them on the RGB video data in our training set. To fine-tune the video model, we use videos snippets that correspond to the RF tracklets in the training set, and to fine tune the image-based model we use the corresponding images in the training set. During testing, we evaluate the performance of RGB-based models on the same query and gallery set we used to evaluate \name. For video-based ReID, the input is the corresponding RGB video of each tracklet. For image-based ReID, we compute the features for each frame in the video, and average them to get the features of the sample.

\begin{table*}[h]
\centering
\begin{tabular}{lcccccccc}\hline
%Method & mAP & CMC-1 & CMC-5 & mAP & CMC-1 & CMC-5 & mAP & CMC-1 & CMC-5\\
\multirow{2}{*}{Methods} & \multirow{2}{*}{Modality} & \multicolumn{3}{c}{RRD-Campus} & \multicolumn{3}{c}{RRD-Home} &\\ \cmidrule(lr){3-5} \cmidrule(lr){6-8} &  & mAP & CMC-1 & CMC-5 & mAP & CMC-1 & CMC-5  \\
\hline
Luo \textit{et al.} \cite{luo2019bag} & RGB Image & 41.3  & 61.4  & 84.3 & - & - & -  \\
Gao \textit{et al.} \cite{gao2018revisiting} & RGB Video & 48.1 & 69.2  & 89.1 & - & - & - \\   
RF-ReID (separate) & RF Signal & 59.5  & 82.1  & 95.5 & 46.4 & 74.6 & 89.5 \\
RF-ReID (combined) & RF Signal & \textbf{60.7}  & \textbf{83.6}  & \textbf{96.5} & \textbf{49.4} & \textbf{75.8} & \textbf{92.5}\\
\hline 
\end{tabular}
\caption{\footnotesize{Comparison between RF-ReID and RGB-based and Video-based ReID on RRD-Campus and RRD-Home. RF-ReID (separate) is trained and tested on RRD-Campus and RRD-Home separately. RF-ReID (combined) is trained on both RRD-Campus and RRD-Home (i.e., the RRD dataset) and tested on both of them.}}
\label{tab:result-main}
\vspace{-5pt}
\end{table*}

\subsection{Quantitative Results}
We compare RF-ReID with state-of-the-art image-based and video-based person ReID models on RRD-Campus. As shown in \autoref{tab:result-main}, our RF-ReID model exhibits a significant improvement over both image-based and video-based models. This is mainly because people in RRD-Campus tend to wear different clothes on different days. Traditional RGB-based ReID models focus on extracting features from clothes, and fail when the same person wears different clothes. In contrast, RF-ReID focuses on the shape and walking style of a person, which remain valid over a long period. 
%hence, the generated features are more robust, leading to better performance on our multi-day RDD-Campus dataset. 

We also report the performance of RF-ReID on RRD-Home. Due to privacy reasons, this dataset does not include RGB images or videos and hence we cannot compare with RGB-based baselines.  In contrast, since RF signal is privacy preserving, it is used to track people in their homes. The results from RDD-Home in \autoref{tab:result-main} show that our model not only achieves high accuracy on RRD-Campus, but also performs well in real-world home scenarios. Furthermore, since humans cannot recognize a person from RF signals, our model can be used to ReID people without collecting personal information like images or videos of people performing private activities in their homes.   

The results in \autoref{tab:result-main} highlight the following points:
\begin{Itemize}
\item RF-ReID works well for long term re-identification that spans days, weeks, or longer. 
\item RF-ReID can re-identify people without collecting or exposing any information that can be used by a human to recognize people; we refer to this property as privacy conscious ReID system. This property is critical in healthcare applications and clinical research, where one needs to ReID the subjects so that one may track changes in a patient's health over time or as a result of treatment; yet, it is essential to keep the subjects de-identified and avoid collecting or storing data that exposes the subjects' personal information.
\item Last, the results indicate that ReID is harder for RDD-Home than RDD-Campus. We believe this is due to two reasons: First, tacklets from homes are shorter (9.2 vs. 11 seconds). Second, the walking style of Parkinson's patients may differ through the day as the impact of medications wears off and they need to take the next dose.  
\end{Itemize}

\subsection{Ablation Study}
We conduct several ablation studies to evaluate the contribution of each component of RF-ReID. All ablation results are for RRD-Campus.

\vskip 0.05in
\textbf{Multi-task Learning:}
Traditional ReID methods use only triplet loss and ID classification loss. In our RF-ReID model, we have added an additional skeleton loss both for regularization and  to learn human-related information. We evaluate the performance of this skeleton loss. 
\autoref{tab:result-skeleton} shows that adding the skeleton loss improves RF-ReID's accuracy. We also report the commonly used metric: Mean Per Joint Position Error (MPJPE)~\cite{ionescu2013human3,zhao2019through}, to evaluate the accuracy of the generated skeletons. As shown in the last column of \autoref{tab:result-skeleton}, the features from RF-ReID contain enough information to generate accurate human skeletons.

\begin{table}[h]
\centering
\begin{tabular}{l |@{\hspace{0.2cm}} c @{\hspace{0.2cm}} c @{\hspace{0.2cm}}c @{\hspace{0.2cm}}c}\hline
Method                           & mAP & CMC-1 & CMC-5 & MPJPE \\
\hline
w/o skl loss  & 57.3  & 78.2  & 94.4   & -\\
w/ skl loss & \textbf{59.5}  & \textbf{82.1}  & \textbf{95.5}  & \textbf{7.44}\\
\hline     
\end{tabular}
\caption{\footnotesize{Performance of RF-ReID with and without skeleton loss.}}
\label{tab:result-skeleton}
\vspace{-10pt}
\end{table}

\vskip 0.05in
\textbf{Hierarchical Attention Module:}
RF-ReID has a hierarchical attention module to aggregate features across a tracklet. The attention module has two blocks: the first aggregates shape information within each segment (3 secs), while the second aggregates walking-style information across the whole tracklet. \autoref{tab:result-attention} demonstrates the effectiveness of each block. If both blocks are replaced by average pooling over the temporal dimension, the performance would drop by 4.1\% for mAP and 7.6\% for CMC-1. Each block  increases the mAP by 3$\sim$4\%, and adding them together achieves the highest performance.

\begin{table}[h]
\centering
\begin{tabular}{l |@{\hspace{0.3cm}} c @{\hspace{0.3cm}} c @{\hspace{0.3cm}}c}\hline
Method                           & mAP & CMC-1 & CMC-5  \\
\hline
Avg Pool+Avg Pool & 55.4  & 74.5  & 93.8     \\
Avg Pool+2nd Att. & 58.3  & 80.3  & 94.4    \\
1st Att.+Avg Pool & 58.6  & 81.0  & 94.3    \\
1st Att.+2nd Att & \textbf{59.5}  & \textbf{82.1}  & \textbf{95.5}    \\
\hline     
\end{tabular}
\caption{\footnotesize{Performance of RF-ReID with and without the attention module. The first attention layer (1st Att.) denotes the layer operating within each segment (3 sec), and the second attention layer (2nd Att.) denotes the layer operating on the whole tracklet.}}
\label{tab:result-attention}
\vspace{-10pt}
\end{table}

\vskip 0.05in
\textbf{Environment Discriminator:}
RF-ReID use a discriminator to prevent the model from generating environment dependent features. We evaluate RF-ReID's performance with and without the discriminator. 
%We also evaluate the discriminator's accuracy on environment classification. 
As shown in \autoref{tab:result-dis}, adding the discriminator helps the model improve the performance by a large margin. 

\begin{figure}[h]
\begin{center}
\includegraphics[width=1.0\linewidth]{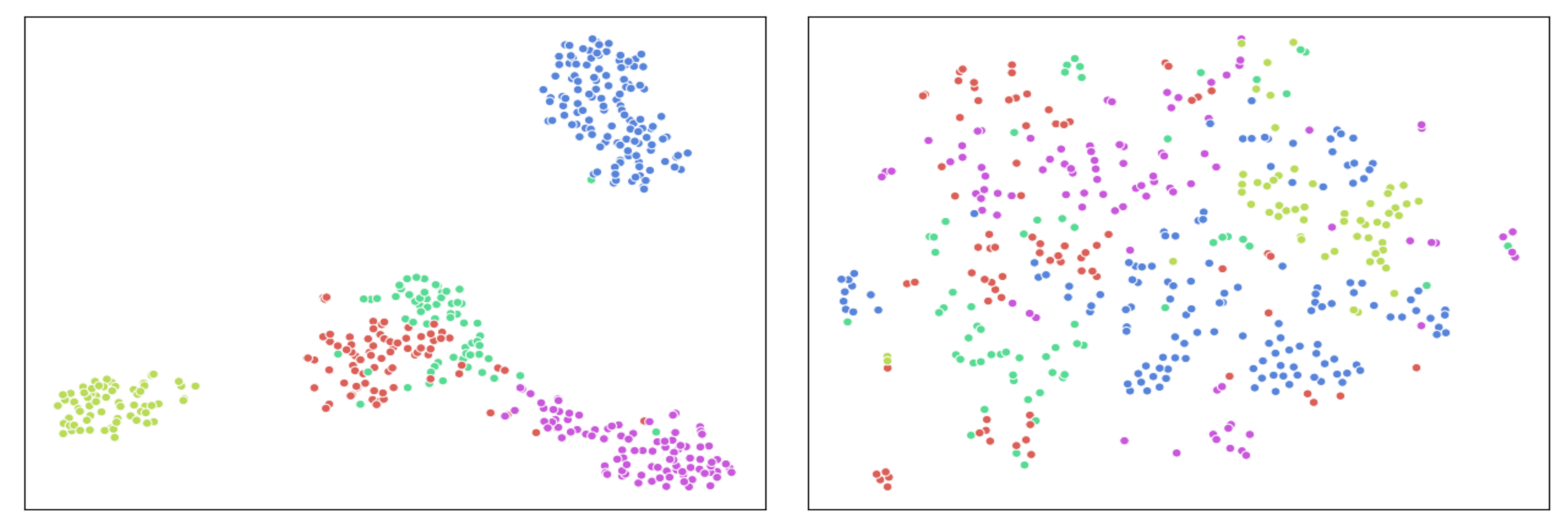}
\end{center}
\vspace{-10pt}
\caption{\footnotesize{Distribution of features extracted by RF-ReID from different environments. The color indicates which environment a feature belongs to. The sub-figure on the left shows the feature distribution without the environment discriminator, where features from same environment are strongly clustered due to environment-dependent information. The sub-figure on the right shows the feature distribution with the environment discriminator. Here, the features are more uniformly spaced, showing that they are more environment-invariant.}}\label{fig:distribution}
\vspace{-10pt}
\end{figure}

\autoref{fig:distribution} visualizes the feature space learned by RF-ReID using t-SNE \cite{maaten2008visualizing}. Each point in the figure corresponds to a feature vector extracted from a tracklet in the test set. There are in total 5 environments in RRD-Campus test set, and we color each point according to its environment.  The figure shows that without the discriminator, the feature distribution is strongly correlated with the environment. In contrast,  with the discriminator, the features are successfully decoupled from the environment and more uniformly distributed. This result further demonstrates that the proposed environment discriminator can help the model learn identifying features focused on the person rather than the environment.

\begin{table}[h]
\centering
\begin{tabular}{l |@{\hspace{0.3cm}} c @{\hspace{0.3cm}} c @{\hspace{0.3cm}}c }\hline
Method                           & mAP & CMC-1 & CMC-5   \\
\hline
w/o discriminator & 56.7  & 74.2  & 93.3   \\
w/ discriminator & \textbf{59.5}  & \textbf{82.1}  & \textbf{95.5}   \\
\hline     
\end{tabular}
\caption{\footnotesize{Performance of RF-ReID with and without the environment discriminator.}}
\label{tab:result-dis}
\vspace{-15pt}
\end{table}

%\begin{table}[h]
%\centering
%\begin{tabular}{l |@{\hspace{0.3cm}} c @{\hspace{0.3cm}} c @{\hspace{0.3cm}}c @{\hspace{0.3cm}}c}\hline
%Method                           & mAP & CMC-1 & CMC-5 & Dis-Acc (\%)  \\
%\hline
%w/o dis loss & 58.3  & 75.0  & 95.3  & 91.7 \\
%w/ dis loss & \textbf{59.8}  & \textbf{82.3}  & \textbf{96.5} & 29.1  \\
%\hline     
%\end{tabular}
%\caption{\footnotesize{Performance of RF-ReID with or without adding the discriminator loss to train the feature generation network.}}
%\label{tab:result-dis}
%\vspace{-15pt}
%\end{table}
\begin{figure*}[t]
\begin{center}
\includegraphics[width=1.0\linewidth]{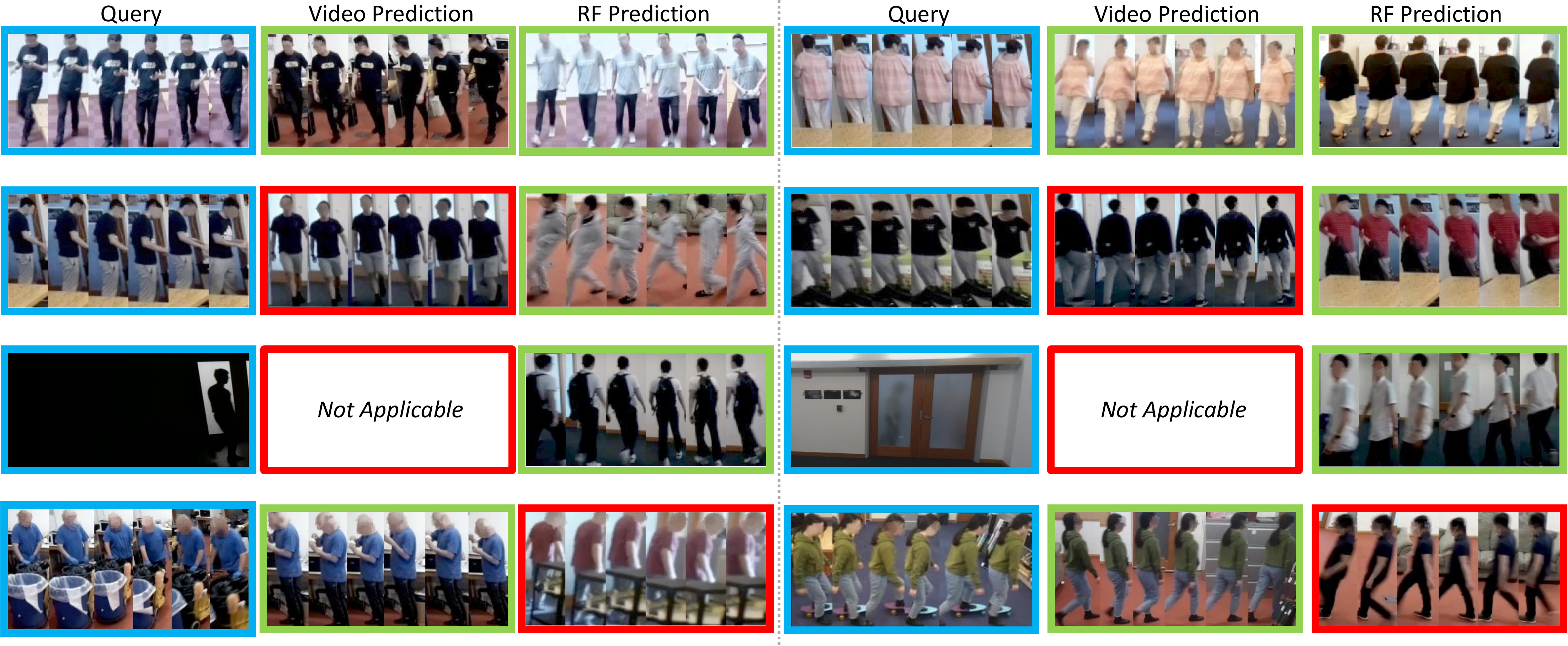}
\end{center}
\vspace{-10pt}
\caption{\footnotesize{Qualitative results on RRD-Campus test set. Each row shows two examples, separated by the dotted line. The first column of each example is a query sample in RRD-Campus test set. The second column is the top-1 prediction by the video-based ReID model in the gallery set. The third column is the top-1 prediction by RF-ReID in the gallery set. Blue boxes stand for query sample. Green boxes mean the prediction is correct, and red boxes mean the prediction is wrong. The first row shows scenarios where both video-based ReID and RF-ReID succeed matching the correct person. The second row shows scenarios where video-based ReID fails, and matches to the wrong person because he has similar clothes, while RF-ReID provides accurate predictions. The third row shows RGB-based ReID fails under dark and occluded conditions but RF-ReID can still work. The last row shows the limitations of RF-ReID which emphasizes the walking style of the person and can get confused when he drags a bicycle or is skateboarding.}}
\label{fig:results}
\vspace{-10pt}
\end{figure*}

\subsection{Qualitative Results}
In \autoref{fig:results}, we show examples from the test set of RRD-Campus. Each example corresponds to a query sample and its closest sample in the gallery. We compare the results generated by RF-ReID to the video-based baseline. 

The figure shows that RGB-based ReID is focused on colors and clothes, where as RF-ReID is resilient to changes in colors and clothing items. In particular, in the second row in the figure, RGB-based ReID fails because the gallery contains other people wearing clothes similar to the query sample. In contrast, RF-ReID identifies the correct person even if he/she wears completely different clothes in the gallery. 
This demonstrates the robustness of RF-ReID against changes in clothes. 

Further, the third row in the figure shows that RGB-based ReID fails when faced with poor lighting or occlusions. In the example on the left, the light is turned off, and as a result the RGB-based ReID model fails to detect the person altogether, while the RF-based model works accurately without being affected by poor lighting. Additionally, in the example on the right, the person is behind a door so the camera can only see a vague shadow. RF-ReID can still work in this scenario because RF signals naturally traverse walls and occlusions.

We also observe that RF-based ReID can fail under some circumstances, as shown in the last row. In the example on the left, the person in the query is walking with a bicycle. The bicycle changes the person's walking style; it also disturb the RF reflections due to its metallic frame, leading to inaccurate prediction. Another failure case is when the person is on a skateboard. Since RF-ReID focuses on the person's walking style, it fails to identify this person correctly.

\section{Conclusion}
We have proposed RF-ReID, a novel approach for person ReID using RF signals. Our approach can extract longterm identifying features from RF signals and thus enables person re-identification across days, weeks, etc.  This is in contrast to RGB-based ReID methods which tend to focus on short-lived features such as clothes, bags, hair style, etc.  Also, unlike cameras, RF signals do not reveal private or personal information. Thus, our ReID method can be used in healthcare applications and clinical studies where it is important to track the health of each subject over time, while keeping the data de-identified. Finally, RF signals work in the presence of occlusions and poor lighting conditions, allowing us to ReID people in such scenarios.  We believe this work paves the way for many new applications of person ReID ,where it is desirable to track people for a relatively long period and without having to collect their personal information. 

\clearpage
{\small
\bibliographystyle{ieee_fullname}
\bibliography{reference.bib}
}

\end{document}